\documentclass[letterpaper, 10 pt, conference]{ieeeconf}  

\IEEEoverridecommandlockouts                              

\usepackage{graphicx}
\usepackage{amsmath}
\usepackage{multirow}
\usepackage{adjustbox}

\title{\LARGE \bf
Template Matching Advances and Applications in Image Analysis  
}
 
\author{Nazanin Sadat Hashemi $^{1,*}$, Roya Babaei Aghdam $^{1}$, Atieh Sadat Bayat Ghiasi $^{1}$, Parastoo Fatemi $^{1}$
\thanks{$^{1}$ Islamic Azad University, North Tehran Branch, Faculty of Engineering, Department of Information Technology}
\thanks{$^{*}$ Corresponding Author: Nazanin Sadat Hashemi}
\thanks{\:\: \tt\small nazaninhshmi@yahoo.com}
\thanks{}
\thanks{\textbf{Keywords: \textit{Template Matching, Image Processing, Image Analysis} } 
}}

\begin{document}

\maketitle
\thispagestyle{empty}
\pagestyle{empty}

\begin{abstract}

In most computer vision and image analysis problems, it is necessary to define a similarity measure between two or more different objects or images. Template matching is a classic and fundamental method used to score similarities between objects using certain mathematical algorithms. In this paper, we reviewed the basic concept of matching, as well as advances in template matching and applications such as invariant features or novel applications in medical image analysis. Additionally, deformable models and templates originating from classic template matching were discussed. These models have broad applications in image registration, and they are a fundamental aspect of novel machine vision or deep learning algorithms, such as convolutional neural networks (CNN), which perform shift and scale invariant functions followed by classification. In general, although template matching methods have restrictions which limit their application, they are recommended for use with other object recognition methods as pre- or post-processing steps. Combining a template matching technique such as normalized cross-correlation or dice coefficient with a robust decision-making algorithm yields a significant improvement in the accuracy rate for object detection and recognition.
\end{abstract}

\section{Introduction}
In the majority of object detection problems in computer vision and image processing, it is often necessary to determine a measure of the similarities between different scenes that have been presented to the system. The detection and recognition of objects in images is a key research topic in the computer vision community \cite{cox1995template}. A very common approach to detecting objects and finding the similarity measurement is template matching. Template matching tries to answer one of the most basic questions about an image: if there is a certain object in a given image, and where it is found \cite{perveen2013overview}. The template is a description of that object (and hence is an image itself), and it is used to search the image by computing a difference measure between the template and all possible portions of the image that could match the template. If any of these steps produces a small difference, then it is viewed as a possible occurrence of the object. Template matching is a high-level machine vision technique that allows for the identification of those parts of an image that match the given image pattern. Some of its wide-spread applications, including matching object to location and edge detection of images, may be employed to plot a route for mobile robots and in image registration techniques, which also have applications in medical imaging. In other words, image matching is a fundamental aspect of many issues in computer vision, including object or scene recognition, solving for 3D structure from multiple images, stereo correspondence, and motion tracking. The distinction between different matching primitives is probably the most prominent difference between the various matching algorithms. Template matching is a technique for finding areas of an image that match (or are similar) to a template image which requires two images. Source image (I): The image in which we expect to find a match to the template image. Template image (T): The patch image which will be compared to the template image. The ultimate goal is to detect the best technique for the highest matching area \cite{atallah2001faster} \cite{singh2012review} \cite{mahalakshmi2012review}.

\section{Methods and Mathematics}
Measures of match and template matching in two or three dimensional images require a measure of match between two images that indicates the degree of similarity or dissimilarity between them \cite{cox1995template}. Matching algorithm selection depends on target and template images. On particular method categorizes template or image matching approaches into template- or area-based approaches and feature-based approaches. 
\subsection{Featured-based Approach}
A featured-based approach is appropriate when both reference and template images contain more correspondence with respect to features and control points. In this case, features include points, curves, or a surface model to perform template matching. In this category, the final goal is to locate the pair-wise connections between the target or so-called reference and the template image using spatial relations or features. In this approach, spatial relations, invariant descriptors, pyramids, wavelets and relaxation methods play an important role in extracting matching measures.
\subsection{Area-based Approach}
Area-based methods, which are usually known as correlation methods or template matching, were developed for the first time by Fonseca et al. \cite{fonseca1996registration} and are based on a combined algorithm of feature detection and feature matching. This method functions very well when the templates have no strong features with an image, since they operate directly on the pixel values. Matches are measured using the intensity values of both image and template. The matching scores are extracted by calculating squared differences in fixed intensities, correction-based methods, optimization methods and mutual information \cite{mahalakshmi2012review}. In some template matching problems, direct matching between template and target images is impossible. Therefore, the eigenvalue and eigenspace of given images are utilized in template matching. These values provide the details needed to match images under various conditions, such as illumination, colour contrast or adequate matching poses. For instance, a specimen in a given image is searched; the eigenspace may consist of templates of the specimen in different positions to a camera using various lighting conditions or expressions. It is possible for the matching image to be occluded by an object or problems involved in motion and become ambiguous. One of the possible solutions for this is to split the template into multiple sub-images and perform matching on these sub-images.
\subsection{Na\"{\i}ve Template Matching}
\begin{figure}[h!]
	\includegraphics[width=\linewidth]{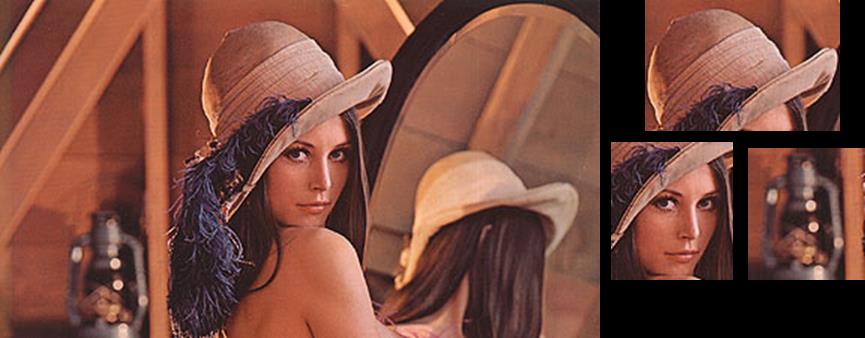}
	\caption{Left image: The Lenna (or Lena) photo is the name given to a standard test image widely used in the field of image processing since 1973. It is a picture of Lena Söderberg, shot by photographer Dwight Hooker. The three images at right are cropped regions, and they are considered templates.}
	\label{fig1}
\end{figure}

Naïve template matching is one of the basic methods of extracting a given which is identical to the template from the image target. In this approach, with or without scaling (usually without scaling), the target image is scanned by the template, and the similarity measures are calculated. Finally, the positions with the strongest similarities are identified as potential pattern positions. Figure \ref{fig1} illustrates the famous image of Lena. As shown, the sub-images are considered as templates, and the naïve matching method is performed in order to extract the templates and patterns from the target images. Because sub-images are originally from the target image without any scaling or manipulation, this straightforward matching algorithm works efficiently.  One of the error metrics which is used to calculate the differences between target and template images is the sum of squared differences \ref{eq1}.
\begin{equation}
\label{eq1}
SSD = \sum_{x,y}^{} [f(x,y)-t(x-u,y-v)]^{2}
\end{equation}
\\
In intensity difference measures, root mean square distance (RMS), sum of the absolute values differences (SAVD), variance of the difference (VOD), variance of the square differences (VOSD) and variance of the absolute valued differences (VOAD) can also be measured depending upon the image’s intensity distribution. 
\subsection{Image Correlation Matching}
In this classic template matching method, the similarity metric between the target and the template is measured. Unlike the naïve template matching algorithm, the target and the template might have different image intensities or noise levels. However, those images must be aligned. The similarity metric used in this approach is based on the correlation between the target and the template.
\subsubsection{Cross-Correlation}
In image processing, cross-correlation is a measure of the similarity of two images where the images are of different sizes. By sliding the first image (template) over the second image (target), the correlation between the two images is measured. The cross-correlation method is similar in nature to the convolution of two functions. Additionally, cross-correlation of a given signal or image with itself is called auto-correlation \cite{perveen2013overview}.
\subsubsection{Normalized Cross-Correlation}
Normalized cross-correlation is an updated version of the cross-correlation approach that has been improved for the following two reasons:
\begin{enumerate}
	\item The results of normalized cross-correlation are invariant to the global brightness changes, and theoretically, the image intensity variations have less effect on this metric compared to the classic method.
	\item The correlation value obtained for a given sliding window is normalized to $[-1, +1]$ interval. Therefore, the normalized cross-correlation values for two similar images trend to +1, and those values for two completely different images trend to $-1$. 
\end{enumerate}
Normalized cross-correlation (NCC) is widely used as an operational resemblance measure for template matching problems. NCC is invariant to linear brightness and contrast variations, and its straightforward implementation has expanded its usage for real-time applications. However, this method is still sensitive to rotation and scale changes, which cause some limitations for deformable templates \cite{cox1995template} \cite{perveen2013overview} \cite{mahalakshmi2012review}.

\begin{equation}
\label{eq2}
NCC = \frac{ \sum_{x,y}^{} [f(x,y)- \overline{f}_{u,v}][t(x-u,y-v)-\overline{t}]}{\sum_{x,y}^{} [f(x,y)- \overline{f}_{u,v}]^{2} \sum_{x,y}^{} [(x-u,y-v)-\overline{t}^{2}]^{0.5}} 
\end{equation}
Equation \ref{eq2} shows the normalized cross-correlation, where $(𝑥)$ is the original image; $\overline{f}$ is the mean of image intensity in the region under the template; $𝑡$ is the template; and $\overline{t}$ represents the mean of image intensity in the template. Here $(𝑥)$ 𝑎𝑛𝑑 $(𝑢)$ represent pixel coordinates \cite{sarraf2016robust} \cite{sarraf2014brain}.
\subsubsection{Matching Filters}
Extensive search template matching methods can be explained as a matching filter of sorts, where template $f(x,y)$ must be detected in a given area of $s(x,y)$. This filter is developed to produce a maximal response where a region of the search area matches the template. 
\begin{equation}
\label{eq3}
z(x,y)=s(x,y)*h(x,y)
\end{equation}
\subsection{Sequential Similarity Detection Algorithms}
Sequential similarity detection algorithms (SSDAs) are a more efficient alternative to correlation-based methods, including matched filters for translational registration. The measure of match is indirectly calculated based on an error for corresponding pixels in f and g in the images under comparison at any stage of the registration process \cite{cox1995template}. 
\subsection{Distortion Measures}
These measures are similar to distortion measures used to evaluate the performance of image compression or image reconstruction techniques. They are also similar to fractal image compression or vector quantisation (VQ), as image matching is used to specify the fractal mappings and to select the VQ codebook. A function space model to derive a reconstruction filter is optimal where the original and reconstructed images are as similar as possible with respect to a particular similarity measure \cite{oakley1990function}. 
\subsection{Dice Coefficient}
Dice Coefficient (DC) has been commonly used in fMRI studies to evaluate the similarity of two spatial maps. Volume-overlap is a simple measure to assess how many of the supra threshold voxels from two $t$ maps occur in the same location \cite{sarraf2016robust} \cite{gorgolewski2013single}. Depending on the normalization factor, there are different variants of the overlap metric; the most common is the Dice Coefficient shown below.
\begin{equation}
\label{eq4}
 DC=\frac{2 \times |A \cap B|}{|A|+|B|} 
\end{equation}
\subsection{Efficient Template Matching Methods}
The algorithm complexity of each template matching algorithm should be calculated. This calculation explains the efficiency of the method in terms of time. Table \ref{table1} demonstrates the computational complexity of the measures of match for certain algorithms. The complexity provided below is defined in terms of the number of operations during calculation of match metric (i.e. cross-correlation) between two images with $r$ rows and $c$ columns. In order to compare the methods, we assumed the template $f$ and image $g$ are digital (digitized) $r×c$ images. The given image $g$ is extracted from each position in a search area called $s$, which has $R$ rows and $C$ columns. The template matching is performed by means of an exhaustive search and translation of the template to each $(R-r+1)×(C-c+1)$ position in the image. Next, a measure of match between the template and the sample at the position of overlap is calculated. The number of operations required for a given template matching method was calculated by multiplying the numbers of operations required by the size of the search area. However, it is possible to increase efficiency because the template does not change during template matching. It is also possible to use redundancy in the calculation of terms for adjacent measures of match to improve efficiency \cite{cox1995template}. 

\begin{table*}[t!]
	\centering
	\caption{Comparison of template matching algorithms for computational complexity based on measure of match.}
	\label{table1}
	\begin{adjustbox}{width=1\textwidth}
		
	\begin{tabular}{|l|l|l|l|l|l|}
		\hline
		\multirow{2}{*}{Measure of Match} & \multicolumn{5}{c|}{Operations}                                      \\ \cline{2-6} 
		& Multiplication & Division & Addition      & Subtraction & Other      \\ \hline
		Normalized Cross-Correlation      & 3rc+1          & 1        & 3(rc-1)       & 0           & 1 sqrt     \\ \cline{2-6} 
		Cross Covariance                  & 3rc+8          & 1        & 5rc-3         & 1           & 1 sqrt     \\ \hline
		Root Mean Square                  & rc+1           & 0        & rc-1          & rc          & 1 sqrt     \\ \cline{2-6} 
		Sum Absolute Valued Diff’s        & 0              & 0        & rc-1          & rc          & rc abs     \\ \cline{2-6} 
		Variance of the Differences       & rc+2           & 1        & 2(rc-1)       & rc+1        & 0          \\ \hline
		Stochastic Sign Change            & r(c-1)         & 1        & (r-1)(c-2)    & rc          & r(c-1) sgn \\ \cline{2-6} 
		Deterministic Sign Change         & r(c-1)         & 1        & rc+(r-1)(c-2) & rc          & r(c-1) sgn \\ \hline
	\end{tabular}
	\end{adjustbox}
\end{table*}
Additionally, certain researchers have used template matching algorithms in frequency domain by mapping the template and target images to the frequency space using Fast Fourier Transform (FFT) \cite{sarraf2016robust} \cite{sarraf2014brain}. They revealed that cross-correlation methods are more efficient than other techniques in frequency domain. However, the application of template matching in frequency domain is limited, as it does not often have any advantage over spatial domain. 

\subsection{Invariant Features}
In practice, one of the challenges of using any template matching techniques is to choose a measure of match that is invariant to image intensity and scale. To overcome this challenge, some scientists have experimented with different scales of a template as well as some rotated version of a template. However, it can only solve the scalability issue of template matching. The image intensity variations remain. Using local or global image intensity normalization, intensity variations in certain problems can be solved. However, this technique is not expandable to most template matching problems.
\subsubsection{Mean Intensity Level Invariance}
For mean intensity level invariant template matching, a measure of match was developed that is based on the variance of the pixel-by-pixel differences between the template and the sub-image of the target image. Equation 5 describes the mean intensity invariant measure of match. 

\begin{equation}
\label{eq5}
 M_{vod} = \frac{1}{1+  \sigma ^{2} } 
\end{equation}
\begin{equation}
\label{eq6}
  \sigma ^{2} =  \frac{ \sum_{A}^{} [( f_{ij} - g_{ij} )-  \mu_{d} ] }{n-1} 
\end{equation}
\begin{equation}
\label{eq7}
\mu_{d}= \frac{\sum_{A}^{} ( f_{ij} - g_{ij} ) }{n} 
\end{equation}
Where $\sigma ^{2}$ is the variance of the differences (VOD) between the template and target images. Additionally,  $\mu_{d}$ is the mean of the differences between the template and target images. Moreover, f and g are two-dimensional signals (images) \cite{cox1993invariance}. 

\subsubsection{Scale Invariance}
One of the major challenges in the generalization of template matching techniques is to provide a scale invariant operator in order to extract the template from the target image. Based on an assumption that target objects in the image are far apart from each other and the background is uniform, the Fourier Mellin (FM) transform can be utilized as a scale invariant representation or operator of the template and the sub-images at each position in the target image before calculating the measure of match \cite{cox1993invariance}. The Fourier Mellin transform includes four steps, which are as follows:
\begin{enumerate}
	\item The Discrete Fourier Transform (DFT) of the image is calculated. The circular shift property of the DFT is a centre translation invariant representation of the original image. However, this implies that the translation limits are the borders of the images.
	\item The magnitude image reconstructed from the DFT operation is normalised by its value at the origin in order to remove the multiplicative effect of scale changes in the original image.
	\item The axes of the positive quadrant of the normalised image are warped onto a logarithmic scale. This converts scaling in the original image to translation. The continuous two-dimensional signal is considered as $f(x,y)$. Let $ \widetilde{F} (u,v)$ be $F(u,v)$ with logarithmically scaled axes.

\begin{equation}
\label{eq8}
f(ax,bv) \longleftrightarrow  \frac{1}{|ab|}F( \frac{u}{a} , \frac{v}{b} ) 
\end{equation}
\begin{equation}
\label{eq9}
 \widetilde{F}_{u,v}=F(ln u, ln v)
\end{equation}
\begin{equation}
\label{eq10} 
\widetilde{F_{norm}}(u,v)=F_{norm}(ln \frac{u}{a} , ln \frac{v}{b} )
\end{equation}
\begin{equation}
\label{eq11}
F_{norm}(ln u - ln a , ln v - ln b)
\end{equation}
The first equation shows the scaled Fourier transform of the 2D signal. In the second equation, the logarithmic representation of the signal is introduced. By combining the first and second equations, the third equation is obtained. The fourth equation clearly demonstrates a translation version of $F_{norm} (ln u,ln v)$.

\item The DFT magnitude of the logarithmically scaled image is calculated, providing a scale in-variant representation of the original image. The scale variations tolerated by the FM transform are limited by the resolution of the image and the size of the image window. 

\end{enumerate}
\subsubsection{Rotation Invariance}
A rotation and scale invariant representation of the template and the sub-image can be generated by simply preceding the procedure described in the previous section with a polar coordinate transform. The origin is at the centre of the template or sub-image, and this converts rotation to a circular shift. The r axis in the polar coordinate system is then warped onto a logarithmic scale, and scaling becomes translation. The DFT magnitude is then calculated for a rotation and scale invariant representation of the image \cite{choi2002novel} \cite{ullah2004using}. Wechsler and Zimmerman \cite{cox1995template} \cite{cox1993invariance} describe a similar approach in terms of conformal complex log mapping. The image is mapped onto a complex plane, where each pixel is represented mathematically, as shown in Equation \ref{eq12}. The complex log mapping that transforms an image from rectangular to polar exponential coordinates is provided in Equation \ref{eq13}. Rotation and scaling become translation in the transform domain. 

\begin{equation}
\label{eq12}
z=x+jy
\end{equation}
\begin{equation}
\label{eq13}
w=ln(z)=ln(|z|)+j\theta_{z}
\end{equation}
\section{Applications}
Template matching algorithms contribute in various applications, from security-based to biomedical-based projects. Although the application of this technique is restricted because of some limitations in shift and scale invariant feature matching, template matching remains a powerful tool for extracting patterns in the pre- and post-processing steps of image analysis. Fast template matching and some deformable-based matching algorithms provide quick solutions for object recognition, while algorithm complexity is not very high \cite{atallah2001faster}. 
\subsection{Deformable Template Models}
\begin{figure}[h!]
	\includegraphics[width=\linewidth]{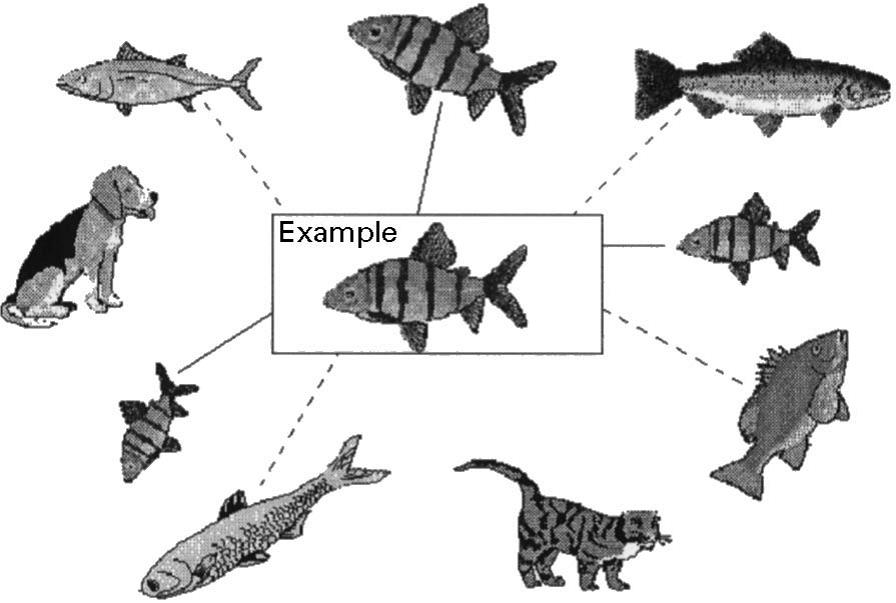}
	\caption{Deformable template matching is shown. The example fish can only be matched to 3 fishes (solid lines) with rigid template matching using translation, scaling and rotation. It can be matched to all of the fishes using deformable template matching (dashed lines).}
	\label{fig2}
\end{figure}

\begin{figure}[h!]
	\includegraphics[width=\linewidth]{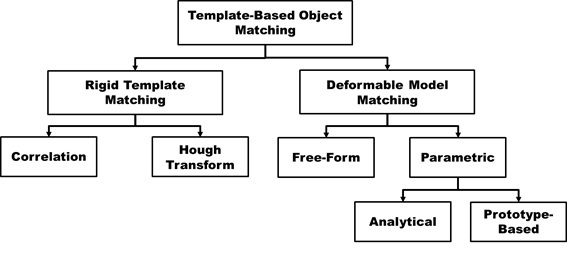}
	\caption{An overview of template matching techniques.}
	\label{fig3}
\end{figure}

In computer vision, model-based shape matching is of interest to many scientists. The first research in this area basically focused on rigid shape matching, where the matched shapes were obtained by applying simple transformations such as translation, rotation, scaling, and affine transformation to the model template, which can be recovered using correlation-based matching or the Hough transform. The concept of deformable templates was simultaneously introduced in computer vision by Widrow \cite{widrow1973rubber} with rubber masks, and by Fischler and Elschlager \cite{fischler1973representation} with spring-loaded templates. Deformable template matching is a more powerful technique because of its ability to address shape deformations and variations, as shown in Figure \ref{fig2}.
Over the past decade, a high volume of research has been conducted on deformable template matching. These activities are divided into two groups. The freeform models represent any arbitrary shape as long as some general regularization constraint (continuity, smoothness, etc.) is satisfied. These restraints are called active contours. On the other hand, parametric deformable models are capable of encoding a specific characteristic shape and its variations. The shape can be characterized by a parametric formula or by using a prototype and deformation modes. An overview of various template matching techniques is provided in Figure \ref{fig3} \cite{jain1998deformable}.
\subsection{Face Detection}
\begin{figure}[h!]
	\includegraphics[width=\linewidth]{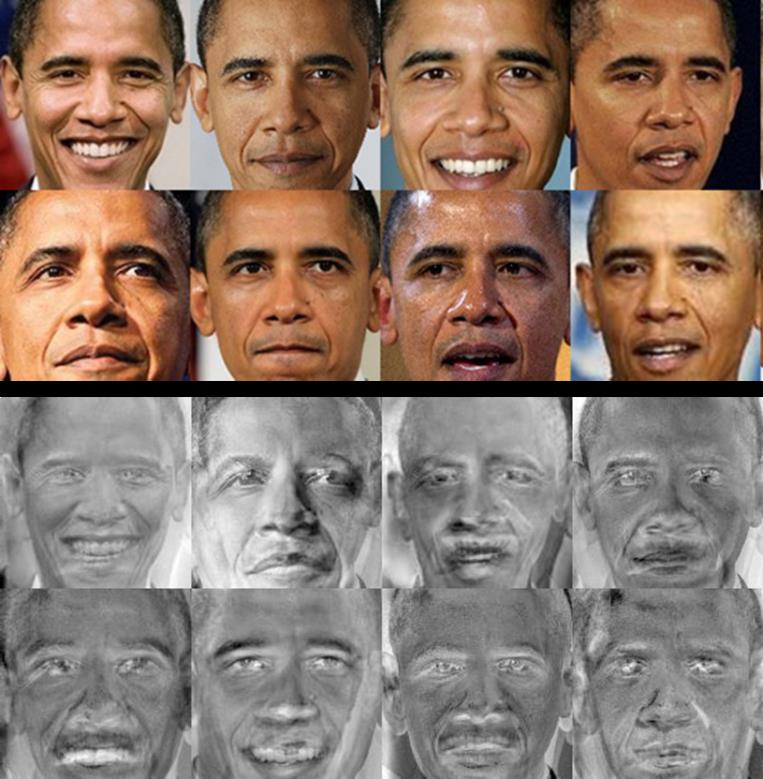}
	\caption{Face detection of President Obama in an online video using combined template matching and eigenvalue decomposition.}
	\label{fig4}
\end{figure}

Face detection is a technique that is used to find an arbitrary sub-image representing the human face of a given and global image. This field of computer vision and image processing is broadly used in security applications, video surveillance, tracking, etc., and it may represent all or only a part of such systems. The traditional method of extracting a given face from a global image is to apply a matching algorithm to the target image. One of the most important factors in face detection is a set of human facial features, which plays a crucial role in the process. Over the past decade, a high volume of research has been performed in the field of face detection, which resulted in the determination that the eyes, mouth and nose are the most important features in both face detection and recognition. Additionally, human faces have a variety of emotions, which are exemplified by many different expressions, and this system can detect the corner of the features in the case of neutral, sad, happy and surprised emotions.  Template matching methods are also applied in facial feature extraction methods, which are sensitive to various non-idealities, such as variations in illumination, noise, orientation, time consumption and color space used. Additionally, strong feature extraction will increase the performance of a face recognition system, as shown in Figure \ref{fig4} \cite{perveen2013overview}.
\subsection{Eye Detection}
\begin{figure}[h!]
	\includegraphics[width=\linewidth]{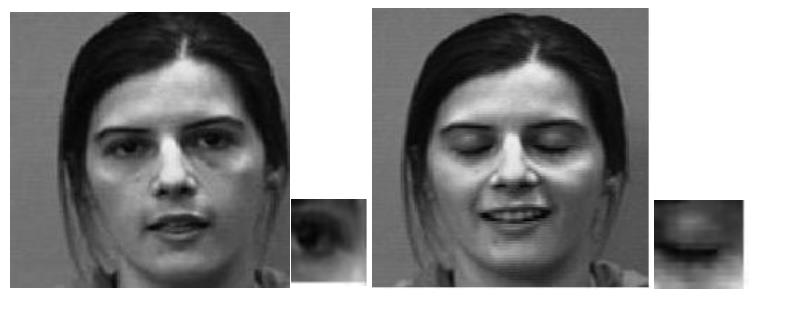}
	\caption{Facial images, including a close-up version of an eye extracted from the face at lower right.}
	\label{fig5}
\end{figure}

Eye detection is a prerequisite for many applications, such as human–computer interfaces, iris recognition, driver drowsiness detection, security, and biology systems. In this paper, template-based eye detection is described. The template is correlated with different regions of the facial image. The region of face which provides the maximum correlation with the template is the eye region. The method is simple and easy to implement. The effectiveness of the method is demonstrated in both open eyes as well as closed eyes through various simulation results, as shown in Figure \ref{fig5} \cite{perveen2013overview}.
\subsection{License Plate Recognition}
\begin{figure}[h!]
	\includegraphics[width=\linewidth]{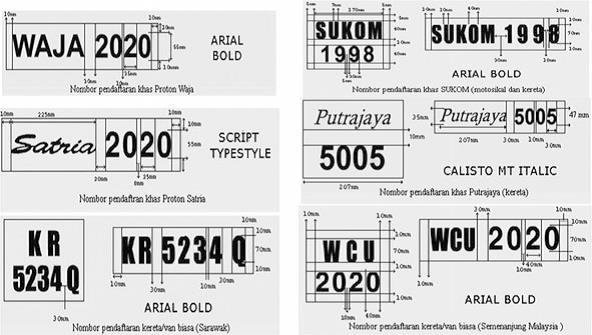}
	\caption{Vehicle number plate specification approved by Jabatan Pengangkutan Jalan Malaysia (JPJ).}
	\label{fig6}
\end{figure}

\begin{figure}[h!]
	\includegraphics[width=\linewidth]{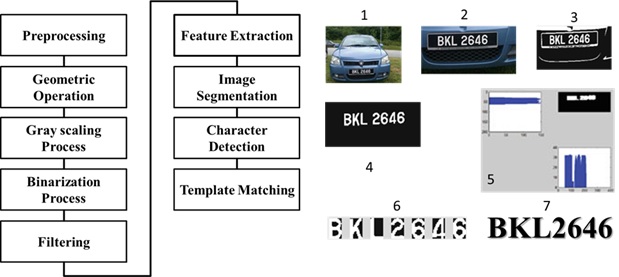}
	\caption{The licence plate detection system used the template matching technique. The first step is preprocessing, which is followed by some geometric operations. Next, gray scaling and binarization are applied to the image. In order to smooth the image, the sample is filtered. In this pipeline, the edge-related features, as well as the horizontal and vertical projection / histogram, are extracted. Then, using all previous information, the licence plate is segmented. Character detection is applied, and finally, using the template matching technique, the characters are recognized.}
	\label{fig7}
\end{figure}

An automatic license plate detection and recognition system is a special form of optical character recognition and has been an active research domain in the image processing field. In a practical sense, template matching is one of the algorithms that can be used to detect licence plates and extract digits or letters from the detected plate. Although the limitations of template matching have been discussed above, the limited styles of licence plates, digits and letters in a given region create the opportunity to use template matching in this popular computer vision problem. The license plate recognition system is a complex image processing application that recognizes the characters on an auto license plate based on the given conditions and situation. The license plate recognition system is installed in many places with multiple purposes, and even law enforcement is using this application to detect speeding vehicles and conduct monitoring and surveillance from a distance. Jalil et al. developed a pipeline to detect and extract Malaysian (Figure \ref{fig6}) licence plates (Figure \ref{fig7}) \cite{jalil2015utilization}.
In the template matching step of the pipeline shown in Figure \ref{fig7}, the size of each character image is normalized according to the template stored in the database. Each character is compared to its corresponding pixel in the template, and the highest coefficient result is identified as the character of the input. This method has demonstrated a high accuracy rate, but it requires an efficient searching method and requires large storage capabilities to save all of the number and character templates.
Template matching is an effective algorithm for recognition of characters used in this pipeline. The character image is compared with those in the database, and the highest correlation value is selected. Correlation is an effective technique for image recognition. This method measure the degree of similarly corresponding pixel values of two images: the template image and the input image. An image with the highest correlation value equates to a strong relationship between the input image and the template image, which will in turn produce the best match. Jalil et al.  showed that by using 100 license plates containing 693 characters for validation of the pipeline, an overall accuracy rate of 92.78 percent for all characters was achieved.

\subsection{Automatic Bare PCB Board Testing}
\begin{figure}[h!]
	\includegraphics[width=\linewidth]{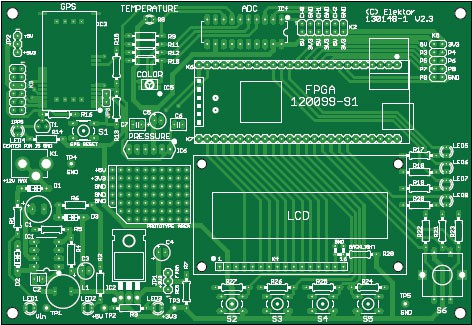}
	\caption{Bare PCB of Multi I/O for FPGA Development Board (130148-71).}
	\label{fig8}
\end{figure}

Another traditional application of template matching is in the detection of loose connections in electrical boards. A bare printed circuit board is a PCB without any placement of electronic components, and it is utilized with other components to produce electric instruments at the circuit level. In order to reduce spending in manufacturing caused by a defective bare PCB, the bare PCB must be inspected. Kanimozhi et al. demonstrated how a template matching-based algorithm is used to check all connections on a given PCB \cite{kanimozhi2013review}. They proposed a novel hybrid approach to the automated inspection of printed circuit boards. A model based on the coordinates and connectivity analysis of the circles was formed using some new approaches to edge linking and the fusion of some edge-based and region-based algorithms. A modified Canny edge detector is used as an edge-based algorithm, while an unsupervised learning algorithm is used to differentiate between regions on the PCB. The researchers also defined a membership function, which fuses the above two results. The edge-linking algorithm extracts the connectivity information for the circles using a new approach that bases decisions on fixation points. In addition, Kanimozhi et al.  developed certain methodologies for locating and identifying multiple objects in images used for surface mounting device inspection. One of the main challenges for surface mounting device inspection is component placement inspection. Component placement errors, which includes missing, misaligned or incorrectly rotated components, are a major cause of defects and must be detected before and after the solder reflow process. The focus of this work was on automated object-recognition techniques for locating multiple objects using grey-model fitting to produce a generalized template for a set of components. The work uses the normalized cross correlation (NCC) template-matching approach and examines a method for constraining the search space to reduce computational calculations. The search for template positions has been performed exhaustively using a genetic algorithm and the experimental results using a typical PCB image (Figure \ref{fig8}) are reported.

\subsection{Normal Breast and Mammogram Positioning}
\begin{figure}[h!]
	\includegraphics[width=\linewidth]{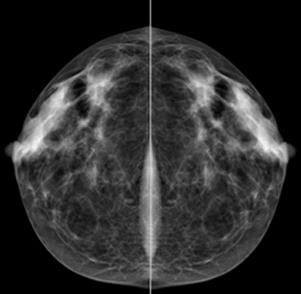}
	\caption{Symmetrical breast mammogram (image courtesy “Medscape”).}
	\label{fig9}
\end{figure}

Over the past several years, the application of template matching in medical image processing and analysis has greatly expanded. Sulaiman et al. \cite{sulaiman2007expert} proposed an expert image processing system based on template matching to locate and determine the symmetrical value between breast mammograms.  They implemented a new template matching algorithm using a cross-correlation method. Their 2D cross-correlation algorithm produced a highly accurate result in the matching process. The correlation method is implemented using the Fast Fourier transform and is modified into a discrete correlation expression for computer usage. Discrete correlation is a mathematical comparison process in which one image is discretely compared to another. The resulting image is a two-dimensional expression of equivalence. In the mammogram viewing process shown in Figure \ref{fig9}, it is important to note the exact orientation of the image. Generally, the breasts are one of the human’s symmetrical organs and are best viewed in symmetric orientation. Comparison of the right breast to the left breast is done for the evaluation of symmetry between both sides. 
In this method, the effectiveness of the proposed template matching system has been evaluated using the sample of breast images. Each sample included the left and right breasts one of each patient. These images are declared symmetrical breasts (normal breasts). For each sample, one side of a breast is established as the template, and another side is established as the search area. The resultant output image is in black and white pixels. The white pixels depict the similarity value, while the black pixels depict the difference value between both images. Therefore, the matching accuracy becomes higher when the white pixels appear more predominantly than the black pixel in the output image. By using the ROI technique, two sample images are created from the original images and are tested for accuracy evaluation of the system. By calculating the white pixels, researchers obtained an understanding of the accuracy of the method, as well as the similarity between the left and right breasts (Figure \ref{fig9}).
\subsection{Brain Network Extraction and Optimization}
\begin{figure}[h!]
	\includegraphics[width=\linewidth]{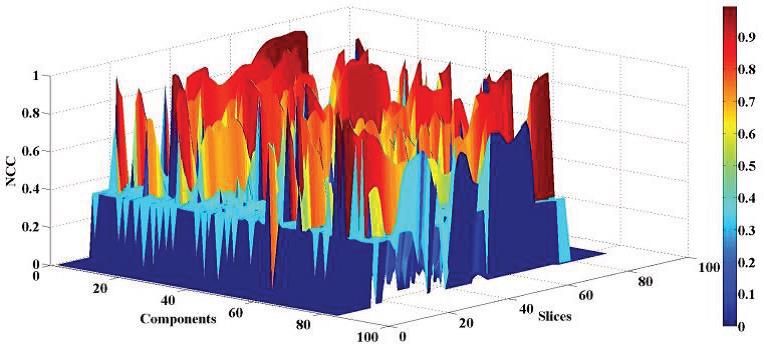}
	\caption{NCC over components and slices (Sarraf et al., 2014, brain network extraction).}
	\label{fig10}
\end{figure}

\begin{figure}[h!]
	\includegraphics[width=\linewidth]{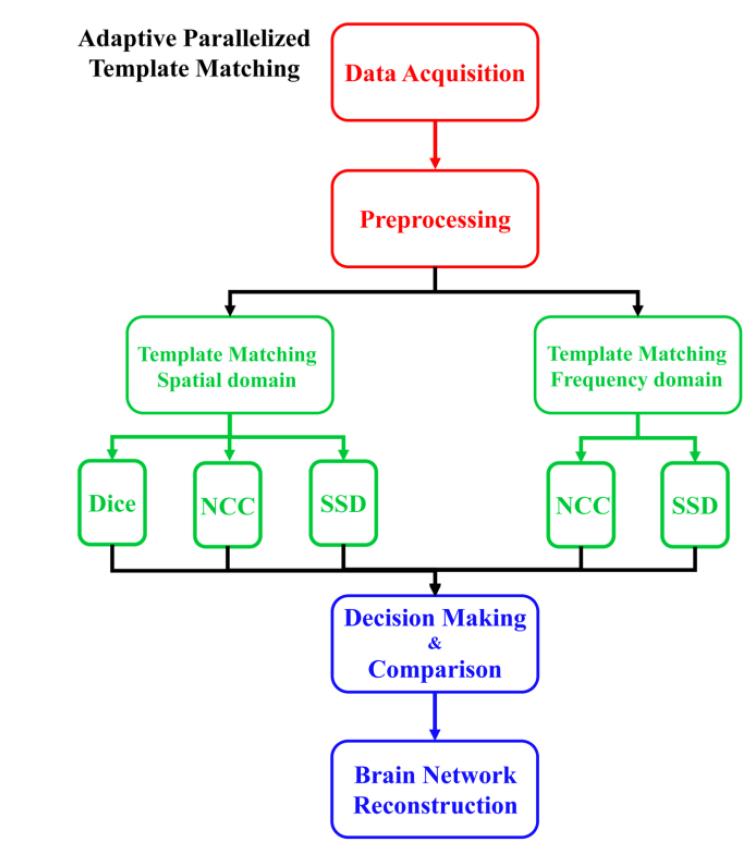}
	\caption{Adaptive and Robust Decision Making Pipeline developed by Sarraf et al. 2016.}
	\label{fig11}
\end{figure}

\begin{figure}[h!]
	\includegraphics[width=\linewidth]{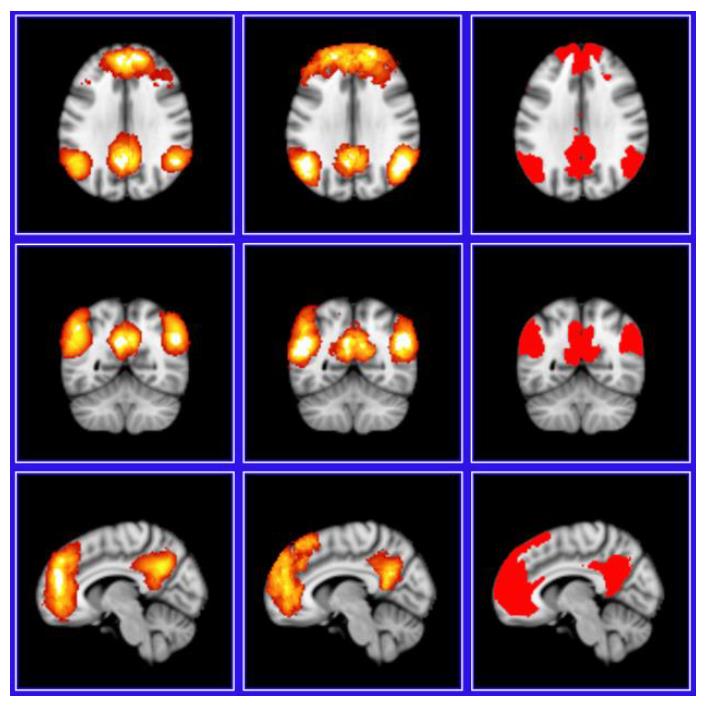}
	\caption{Final DMN mask created by NCC (left), Dice (Middle) and Original DMN template (right). Axial (Top), Coronal (Middle), Sagittal (Bottom) (Sarraf et al.).}
	\label{fig12}
\end{figure}

Recently, researchers have shown great interest in using template matching techniques in medical imaging, with a particular focus on brain imaging. Sarraf et al. \cite{sarraf2016robust} \cite{sarraf2014brain} \cite{Sarraf_2016} used template matching techniques in order to extract a given brain network from magnetic resonance images and also developed a robust decision-making algorithm to optimize the results produced from the matching algorithm. They also experimented with their algorithm in both spatial and frequency domains.  They used the functional magnetic resonance imaging (fMRI) modality, which revealed that certain neural structures are highly active during periods of rest, referred to as a resting-state fMRI. Several methods have been implemented in order to analyze resting-state fMRI data, and probabilistic independent component analysis (PICA) is currently the most popular technique. The major challenge of using PICA is that resting-state networks are split into several components, and visual extraction can be difficult. Sarraf et al. proposed a fast and precise algorithm based on advanced template matching in a spatial domain such as normalized cross correlation (NCC) adapted to functional images to automatically extract the default mode network (DMN), which is the task of the independent resting state network in the brain using PICA. To do so, the researchers created the DMN template and performed template matching. However, they used an arbitrary method to select the best matched network slices (Figure \ref{fig10}). 
In their second approach, Sarraf et al. proposed an adaptive pipeline, as shown in Figure 11, which extracts the brain and network slices as candidates, and by performing a semi-iterative technique that optimizes the given network and constructs this network using a set of candidates. This robust methodology, which is still based on template matching techniques such as NCC, sum of squared differences (SSD) and dice coefficient (DC), enables researchers to compare the results from each method and choose either the algorithm reconstructed network or their favorite method. In this research scenario, after identifying components within the resting state, the decision-making pipeline selected the components within each method that had the highest matching scores to our DMN template. The final decision of selecting the most prototypical DMN components was made by a comparison between methods. This resulted in a DMN mask that was generated by the components chosen by our decision-making algorithm. To evaluate the accuracy of the decision-maker, a cross correlation between each final mask and the template was measured. Results indicated that the normalized cross correlation method, using both the spatial and frequency domains, and the dice coefficient method generated the optimal DMN mask. This demonstrates the utility of our algorithm in providing an objective method for network extraction (Figures \ref{fig11} and  \ref{fig12}) \cite{saverino2016associative} \cite{sarraf2016deep} \cite{sarraf2016deepad}.
The results demonstrated that NCC and dice coefficient methods extracted the most similar components and ultimately created the most accurate DMN mask. Matching based on the spatial domain versus the frequency domain resulted in a better match with the DMN template, suggesting that spatial domain leads to better performance of decision-making pipelines in the identification of brain networks.
\section{Conclusion}
In this work, the mathematical basics of template matching techniques and various matching algorithms were studied. Additionally, a wide range of template matching applications in image processing, computer vision, medical image analysis, and in particular brain imaging were reviewed. We also demonstrated that template matching algorithms, whether in spatial or frequency domains, should be accompanied with a post-classification or decision-making step in order to improve the accuracy of the matching system. We concluded that feature-based matching methods should be applied when the structural information matches rather than the intensity information. Furthermore, area-based approaches should be used if they do not have many prominent details and the characteristic information is apparent (colour/gray than shape/size). Finally, it is worth remembering that a given template and the input images must have either statistical dependence or intensity similarities. In the case of intensity similarities, the correlation methods can also be used. From a geometric point of view, only small amounts of shifts and rotations are allowed. Otherwise, the deformable model or shift and scale invariant features should be used.
\section*{Disclosure statement } 
The authors declare that there is no conflict of interest regarding the publication of this manuscript.

\end{document}